\pdfoutput=1

\documentclass[11pt]{article}

\usepackage[]{acl}

\usepackage{times}
\usepackage{latexsym}
\usepackage{graphicx}
\usepackage{caption}
\usepackage{subcaption}
\usepackage{hyperref}
\usepackage{booktabs}
\usepackage{makecell}
\usepackage{multirow}
\usepackage{amsmath}
\usepackage{xltabular}
\usepackage[T1]{fontenc}
\usepackage{relsize}

\usepackage[utf8]{inputenc}

\usepackage{microtype}

\title{Simpson's Paradox and the Accuracy-Fluency Tradeoff in Translation}

\author{Zheng Wei Lim, Ekaterina Vylomova, Trevor Cohn,%
\thanks{~~Now at Google.} ~\and Charles Kemp \\
        The University of Melbourne \\
        \texttt{ z.lim4@student.unimelb.edu.au} \\ 
        \texttt{\{vylomovae,t.cohn,c.kemp\}@unimelb.edu.au} }

\begin{document}
\newcommand{\bs}[1]{\boldsymbol{#1}}
\newcommand{\p}[1]{$p(\bs{#1})$}
\newcommand{\accuracym}{$\text{accuracy}_M$}
\newcommand{\fluencym}{$\text{fluency}_M$}

\maketitle
\begin{abstract}

A good translation should be faithful to the source and should respect the norms of the target language. We address a theoretical puzzle about the relationship between these objectives. On one hand, intuition and some prior work suggest that accuracy and fluency should trade off against each other, and that capturing every detail of the source can only be achieved at the cost of fluency. On the other hand, quality assessment researchers often suggest that accuracy and fluency are highly correlated and difficult for human raters to distinguish~\cite{callison2007meta}. We show that the tension between these views is an instance of Simpson's paradox, and that accuracy and fluency are positively correlated at the level of the corpus but trade off at the level of individual source segments. We further suggest that the relationship between accuracy and fluency is best evaluated at the segment (or sentence) level, and that the trade off between these dimensions has implications both for assessing translation quality and developing improved MT systems.

\end{abstract}

\section{Introduction}

\begin{table*}
\centering
\small
\setlength{\tabcolsep}{2pt}
\begin{tabular}{clccccc}
\toprule
\multicolumn{2}{l}{Translation}  & accuracy & fluency & \accuracym & \fluencym & $\log p(\bs{y}|\bs{x})$ \\
\midrule

(i) & \makecell[l]{ Ich gab Ihnen eine Rückerstattung des Buches. }
 & 23.0 & 25.0 & -10.81 & -56.0 & -10.31\\[0.1cm]
(ii) &  \makecell[l]{ Ich habe Ihnen eine Rückerstattung des Buches ausgestellt.} & 24.3 & 24.7  & -6.13 & -64.0 & -12.13 \\[0.1cm]	
(iii) & \makecell[l]{Ich stellte Ihnen eine Rückerstattung des Buches aus.}  & 25.0 & 23.0 & -6.44  & 	-70.0 & -14.75 \\




\bottomrule
\end{tabular}
\vspace{-0.1in}
\caption{Translations of ``\textit{I issued you a refund of the book.}" from English to German, which correspond to three of the orange dots in Figure~\ref{fig:simpson}. Human ratings of accuracy and fluency are derived from MQM scores, and  \accuracym\ ($\log p(\bs{x}|\bs{y})$) and \fluencym\ ($\log p(\bs{y})$) are estimated using an NMT model. Option (i) is acceptable but \emph{gab} (past tense of give) is less accurate than the conjugations of \emph{ausstellen} (issue) used in (ii) and (iii).   Option (iii) is the least natural because  \emph{stellte ... aus} (Pr\"{a}teritum tense) is typically used only in formal writing. }
\label{tab:examples}
\end{table*}

No translation can simultaneously satisfy all possible goals, and translation is therefore an art of navigating competing objectives~\cite{darwish08}. Many objectives are discussed in the literature, but two in particular seem especially fundamental. The first is accuracy (also known as fidelity or adequacy), or the goal of preserving the information in the source text (ST). The second is fluency, or the goal of producing target text (TT) that respects the norms of the target language (TL) and is easy for the recipient to process~\cite{kunilovskaya2023translationese}. 

Here we study the relationship between accuracy and fluency and work with two operationalizations of these notions. The first relies on human judgments of accuracy and fluency collected in prior work on translation quality estimation~\cite{castilho2018approaches}. The second relies on probabilities estimated using neural machine translation (NMT) models. Given a source-translation pair $(\bs{x}, \bs{y})$, \p{x|y} corresponds to accuracy, and \p{y} corresponds to fluency~\cite{teich2020translation}. 
\p{x|y} will be low if $\bs{y}$ fails to preserve all of the information in $\bs{x}$, and \p{y} will be low if $\bs{y}$ violates the norms of the target language. To highlight that model estimates \p{x|y} and \p{y} are related to but distinct from human ratings of  accuracy and fluency, we refer to \p{x|y} as \accuracym\ and \p{y} as \fluencym. 

\begin{figure}
    \centering
    \includegraphics[width=\linewidth]{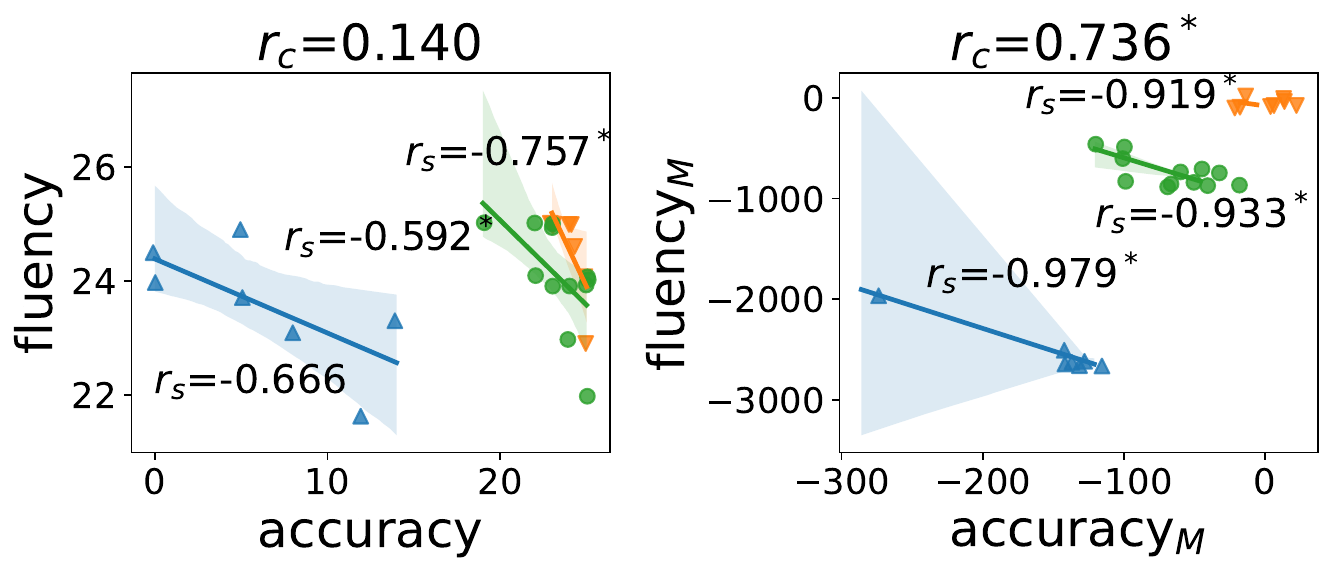}
    \vspace{-0.3in}
     
     \caption{Simpson's paradox. Each panel shows translations of three source segments indexed by color and marker shape. At the source segment level, accuracy and fluency (left) and \accuracym\ and \fluencym\ (right, probabilities plotted on log scales) both show negative correlations $r_s$. 
     At the corpus level, both pairs of dimensions show positive correlations $r_c$ (see panel labels). Significant correlations ($p<.05$) are marked with `*'. Source segments and translations are drawn from past WMT General Task submissions and data points have been jittered for clarity. The shaded areas show 95\% confidence intervals based on 1000 bootstrapped samples. Full  translations are included in Tables~\ref{tab:example1} (orange dots), \ref{tab:example3} (green dots) and \ref{tab:example2} (blue dots) of the appendix.
     }
     \label{fig:simpson}
\end{figure}

Some parts of the literature argue that accuracy trades off with fluency. In Figure~\ref{fig:simpson}a, the blue dots are  translations of the same source segment, and Table~\ref{tab:examples} shows three translations that illustrate the same kind of tradeoff. A translator choosing between these alternatives cannot simultaneously maximize accuracy and fluency, because the most accurate translations are not the most fluent, and vice versa. \citet{teich2020translation} argues that \accuracym\ and \fluencym\ should  trade off in this way, and the same view is implicitly captured by noisy-channel models of translation~\cite{brown1993mathematics}, which aim to generate translations $\bs{y}$ that maximize $p(\bs{y}|\bs{x}) \propto p(\bs{x}|\bs{y})p(\bs{y})$. Typically these models include weights for the two components \p{x|y} and \p{y} that can be interpreted as the extent to which \accuracym\ is prioritized over \fluencym, or vice versa~\cite{yu2016neural,yee2019simple,yu2020better,muller2020domain}. 

An opposing view of the relationship between accuracy and fluency emerges from the literature on quality estimation. Here the common wisdom is that accuracy and fluency are highly correlated and practically indistinguishable to human annotators \cite[but see \protect\citealt{djiako2019lexical}; \protect\citealt{sulem2020semantic}]{callison2007meta,banchs2015adequacy,mathur21}. As a result, accuracy and fluency are conflated as a single assessment score in
recent WMT General Machine Translation Tasks, with more emphasis given to accuracy than fluency \cite{farhad2021findings,kocmi2022findings,kocmi2023findings}.

We argue that the conflict between these views is an instance of Simpson's paradox \cite{yuan2021simpson}, which occurs when a relationship at one level of analysis (e.g.\ the corpus level) disappears or is reversed at a different level (e.g.\ the segment or sentence level). Figure~\ref{fig:simpson} shows how the correlation $r_c$ between accuracy and fluency can be positive over a miniature corpus including translations of three source segments even though the correlation $r_s$ for each individual source segment is negative. Of the two levels of analysis, the segment level is the appropriate level for understanding how humans and machine translation systems should choose among possible translations of a source segment. 
The central goal of our work is therefore to establish that the correlation between accuracy and fluency is negative at the level of individual source segments.

\section{Tradeoff between \p{x|y} and \p{y}}
\label{sec:prob_tradeoff}
Because \accuracym\ and \fluencym\ have formal definitions, we start with these dimensions. 

\subsection{Theoretical formulation and simulation}

Let $\bs{Y}$ be a finite set of translations of source segment $\bs{x}$, and let $\vec{p}_{x|y}$ and $\vec{p}_{y}$ denote log probability vectors that include \accuracym\ and \fluencym\ scores for all $\bs{y} \in \bs{Y}$.\footnote{There are infinitely many possible translations, but here we consider a finite set generated by humans or machines.} We use the Pearson correlation between the two vectors:
\begin{align}
    r_s &= \mathrm{corr}(\vec{p}_{x|y}, \vec{p}_{y}) \label{eq:tradeoff}
\end{align}
to quantify the tradeoff between \accuracym\ and \fluencym\ across  translations of $\bs{x}$.
If $r_s > 0$ there is no tradeoff, and the translations with higher \accuracym\ also tend to have higher \fluencym.  If $r_s < 0$ the dimensions trade off, and improving a translation along one dimension tends to leave it worse along the other. Note that $r_s$ is a correlation at the segment level, and should be distinguished from the corpus-level correlation $r_c$ between \p{x|y} and \p{y} over an entire corpus of segments $\bs{x}$ and their translations $\bs{y}$. 

Suppose that a translator is considering candidate translations $\bs{y}$ of source segment $\bs{x}$. There are a vast number of possible translations, including many nonsense translations, but we assume that the translator chooses among a small set of good translations that all have near-maximal  values of \p{y|x}. 
 Because $p(\bs{y}|\bs{x}) \propto p(\bs{x}|\bs{y})p(\bs{y})$ is roughly constant over this set of good translations, it follows that \accuracym\ and \fluencym\ trade off within the set.  

To validate this informal argument, we ran simulations to confirm that tradeoffs between 
\p{x|y} and \p{y} emerge when $\bs{x}$ and $\bs{y}$ are  numeric vectors drawn from a Gaussian joint distribution  $P(\bs{x}, \bs{y})$ centered at zero.\footnote{Code available at \url{https://github.com/ZhengWeiLim/accuracy-fluency-tradeoff}.}
We set an initial square matrix $A$ with dimensionality equal to the total number of dimensions in $\bs{x}$ and $\bs{y}$ combined. Assuming all elements in $\bs{x}$ and $\bs{y}$ have $\sigma^2=1$ and pairwise positive covariance, all diagonal elements of $A$ are set to 1  and other elements 0.7. To ensure the covariance matrix is positive semi-definite, we replace the initial matrix $A$ with a final covariance matrix defined as $A^\top A$.

For each ``source segment'' $\bs{x}$ considered in our simulation, we generate 10,000 possible ``translations'' $\bs{y}$ by sampling from a distribution $q(\bs{y}) = \prod_i q(y_i)$, where each element $y_i$ of $\bs{y}$ is sampled uniformly within two standard deviations of its mean. We then score each translation and  compute \p{x|y}, \p{y} and \p{y|x} 
using the known joint  $P(\bs{x}, \bs{y})$.

\begin{figure}
    \centering
    \begin{subfigure}[b]{\linewidth}
     \includegraphics[width=\linewidth]{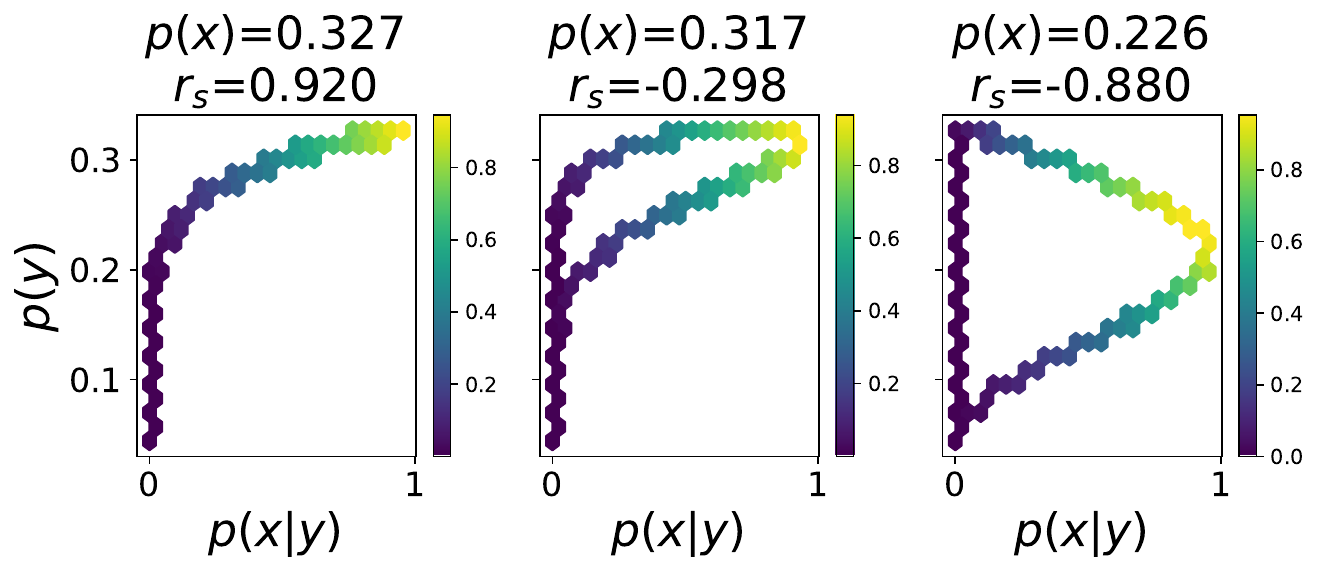}
     \vspace{-0.25in}
     \caption{\textlarger[0.7]{
     Simulation with one-dimensional $\bs{x}$ and $\bs{y}$. The three panels correspond to three different source ``segments'' $\bs{x}$ of decreasing probability \p{x}, and the points in  each panel are candidate translations $\bs{y}$.  Brighter colors indicate translations with larger \p{y|x}. Pearson correlations across translations ranked in the top 10\% based on \p{y|x} are shown at the top of each panel.}}
     \label{fig:px_sim}
     \end{subfigure}
    
    \begin{subfigure}[b]{\linewidth}
    \centering
     \includegraphics[width=\linewidth]{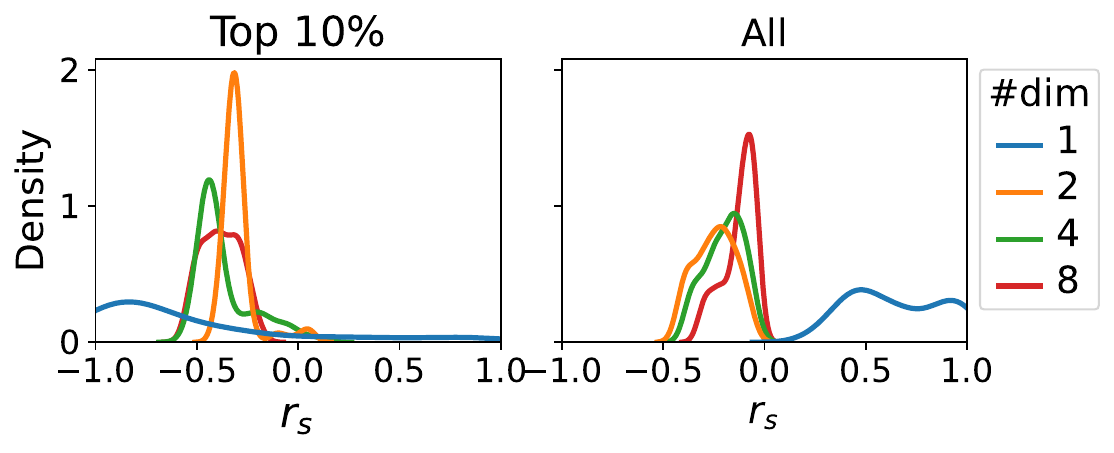}
     \vspace{-0.1in}
     \caption{\textlarger[0.7]{Kernel density plots of tradeoffs across the top 10\% (left) and across all translation choices (right). The tradeoff persists in higher dimensional space, and is stronger when selecting only $\bs{y}$ with the highest values of \p{y|x}. }} 
     \label{fig:dim_sim}
     \end{subfigure}

     \caption{Tradeoffs between \p{x|y} and \p{y} in synthetic data.
     }
\end{figure}

We initially assume that both $\bs{x}$ and $\bs{y}$ are one-dimensional vectors. Figure~\ref{fig:px_sim}
shows the relationship between \p{y} and \p{x|y} for 3 ``segments'' $\bs{x}$. Each point in each panel corresponds to a candidate translation $\bs{y}$, and candidates with highest \p{y|x} are shown in yellow. The correlation above each panel results from applying Equation 1 to all translations with \p{y|x} above the 90th percentile (i.e.\ all points in the brightest part of each plot). The first  ``segment'' $\bs{x}$ (leftmost panel) has relatively high probability \p{x}, and no tradeoff is observed in this case. The tradeoff emerges, however, and becomes increasingly strong as $\bs{x}$ moves away from the mode of the distribution \p{x}. At the ``corpus'' level, 
\p{y} and \p{x|y} are uncorrelated ($r_c=-.001, p=.970$) when the top 10\% of translations for each of the three ``segments'' are combined.

Figure~\ref{fig:dim_sim} shows that the tradeoff persists when the dimensionality of $\bs{x}$ and $\bs{y}$ is increased. The density plot for each dimensionality is based on a sample of 100 source ``segments'' (rather than the 3 in Figure~\ref{fig:px_sim}), and at all dimensionalities the majority of source ``segments'' induce tradeoffs. The tradeoffs are stronger (i.e.\ correlations more negative) when the candidate translations consist of the $\bs{y}$ with highest \p{y|x} (top 10\%), but for all dimensions except $n=1$ most source ``segments'' still induce a tradeoff even if all candidate translations are considered. At the ``corpus'' level, \p{y} and \p{x|y} of the top translations are positively correlated ($r_c=.399, .159, .113, .109$ for dimensionalities 1, 2, 4 and 8 respectively, $p<.001$).

Although our simulations aim for simplicity rather than realism, they  provide theoretical grounds for expecting tradeoffs at the segment level in real translations generated by humans and machines. They also suggest that the tradeoff may become stronger when only high-quality translations are considered, and that the strength of the tradeoff may depend on \p{x}.


\subsection{Human and machine translation}

We now show that human and machine translations show the same tradeoff between \accuracym\ and \fluencym, which correspond to \p{x|y} and \p{y} estimated by an NMT model.

\medskip\noindent
\textbf{Data.} We analyze 15 translation studies from CRITT TPR-DB (CRITT) that include 13 language pairs \cite{carl2016critt}. We also use a subset of the Russian Learner Translator Corpus (RLTC) that has been aligned at the sentence level by \citet{kunilovskaya2023translationese}. For machine translation, we use WMT test sets which include segments of (mostly individual) sentences that are annotated with Multidimensional Quality Metrics labels (MTMQM) \citep{freitag2021experts,freitag2021results,zerva2022findings,freitag2023results}.
To reduce spurious correlations, we remove duplicate translations and source segments with fewer than four unique translations. Additional details are provided in the appendix.

\medskip\noindent
\textbf{Models.} We use NLLB-200's 3.3B variant model \citep{costa2022no} to estimate \p{y|x} and \p{x|y}.\footnote{\href{https://huggingface.co/facebook/nllb-200-3.3B}{NLLB model card}} For consistency, we also extract \p{y} based on the same model, skipping all inputs except for special tokens (e.g., \texttt{<eos>} tags).\footnote{To ensure reproducibility across models, we repeat our analysis in the appendix using M2M100 \cite{fan2021beyond}.} All probabilities are log scaled.

\begin{figure}
         \includegraphics[width=\linewidth]{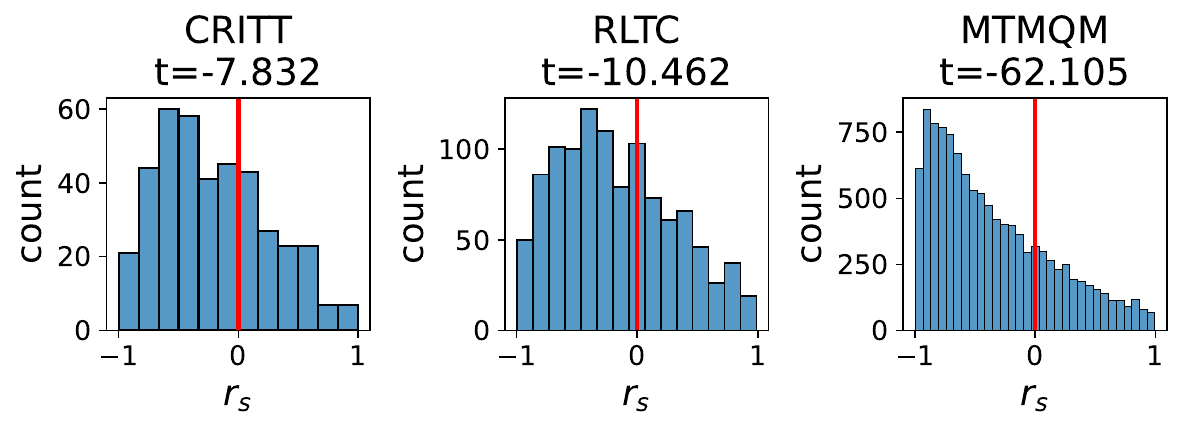}
         \vspace{-0.2in}
         \caption{Tradeoffs between estimated \p{x|y} and \p{y} across source segments from three corpora. Paired-sample t-tests against randomly permuted \p{y} and \p{x|y} are shown at the top of each panel.}
         \label{fig:tradeoff}
\end{figure}

\medskip\noindent
\textbf{Results.} Figure~\ref{fig:tradeoff} is a histogram analogous to the densities in Figure~\ref{fig:dim_sim}, and shows distributions of tradeoff scores for source segments in CRITT, RLTC and MTMQM. In all three cases most source segments induce tradeoffs (i.e.\ produce negative correlations). To test for statistical significance we compared the actual distributions against randomly permuted data. The results of all paired-sample t-tests are significant ($p<.001$), and are included in the figure.\footnote{Each permuted data set is created by randomly shuffling the pairings of \p{x|y} and \p{y} within the set of possible translations of each source segment.} 
When samples are aggregated at the corpus level, \p{y} and \p{x|y} show significant positive correlations ($p<.001$) for CRITT ($r_c=.625$), RLTC ($r_c=.685$) and MTMQM ($r_c=.675$), revealing that Simpson's paradox applies in all three cases.

The simulation in Figure~\ref{fig:px_sim} suggests that segments with smaller \p{x}  tend to show greater tradeoffs, which predicts that  \p{x} and $r_s$ (Equation \ref{eq:tradeoff}) should be positively correlated. Our data support this prediction for CRITT ($r=.124$, $p=.013$), RLTC ($r=.225$, $p <.001$) and MTMQM ($r=.109$, $p<.001$). 

\section{Tradeoff between accuracy and fluency}
\label{sec:rating_tradeoff}

We now turn to human ratings of accuracy and fluency, and demonstrate that the two are again negatively correlated at the segment level. 

\medskip\noindent
\textbf{Data.} Only RLTC and MTMQM are rated by human annotators. The subset of RLTC released by \citet{kunilovskaya2023translationese} includes accuracy and fluency scores derived from error annotations.   For MTMQM, we follow \citet{freitag2021experts} where accuracy scores are aggregates of   ``Accuracy'' and  ``Terminology'' errors, and fluency scores are aggregates of   ``Fluency'', ``Style'' and ``Locale convention'' errors. Targets that are labelled ``Non-translation'' receive scores of zero for both accuracy and fluency. Major and minor errors receive penalties of 5 and 1 respectively. Fluency/Punctuation is assigned a penalty of 0.1. We calculate the final rating as $s_c = \max(0, 25-e_c)$, where $e_c$ denotes the total penalty in error category $c$.\footnote{The maximum score is set at 25 because the maximum MTMQM penalty score is 25.} Because some systems submit the same translation but receive different ratings, we average these scores and remove the duplicate entries.


\begin{figure}
         \begin{center}
         \includegraphics[width=\linewidth]{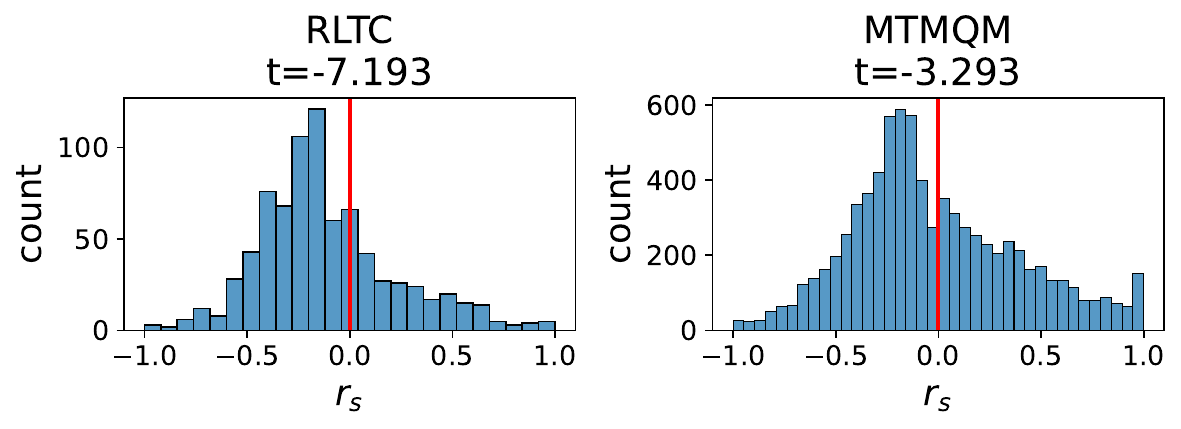}
         \caption{Tradeoffs between human ratings of accuracy and fluency across segments from two corpora. Paired-sample t-tests against randomly permuted scores are shown at the top of each panel.}
         \label{fig:acc-flu-tradeoff}
         \end{center}
\end{figure} 

\medskip\noindent
\textbf{Results.} Figure~\ref{fig:acc-flu-tradeoff} shows correlations at the level of individual source segments. The majority of correlations are negative, and paired-sample t-tests reveal that both distributions are significantly ($p<.001$) different from distributions obtained from random permutations. The results therefore suggest that accuracy and fluency (as rated by humans) trade off at the level of individual segments. At the corpus level, accuracy and fluency are positively correlated for MTMQM ($r_c=.392, p<.001$), and are uncorrelated in RLTC ($r_c=-.085, p<.001$), suggesting again that Simpson's paradox applies to both cases.\footnote{ Fluency and accuracy may be uncorrelated in RLTC at the corpus level because of a ceiling effect -- 63.5\% and 70.6\% of sentences receive maximum ratings for fluency and accuracy in RLTC compared to 55.6\% and 58.4\% for MTMQM. 
}

 Unlike the case for
 \accuracym\ and \fluencym, human ratings of accuracy and fluency do not induce a positive correlation between
 \p{x} and $r_s$ ($r=-.150$ and $-.104$ for RLTC and MTMQM respectively). We therefore find no support for the simulation-based prediction that low-probability sentences are more likely to produce strong tradeoffs between accuracy and fluency. 

Figure~\ref{fig:acc-flu-tradeoff} is directly analogous to Figure~\ref{fig:tradeoff}, and we expected that source segments which showed strong tradeoffs (i.e.\ extreme negative correlations) in Figure~\ref{fig:tradeoff} would also show strong tradeoffs in Figure~\ref{fig:acc-flu-tradeoff}. The two tradeoff measures, however, were  uncorrelated,\footnote{The Pearson correlations between the two tradeoff measures for RLTC and MTMQM are $r=.003, p=.933$ and $r=.022, p=.05$.} which suggests that \accuracym\ and \fluencym\  overlap only partially with human ratings of accuracy and fluency.

\begin{figure}
    \includegraphics[width=\linewidth]{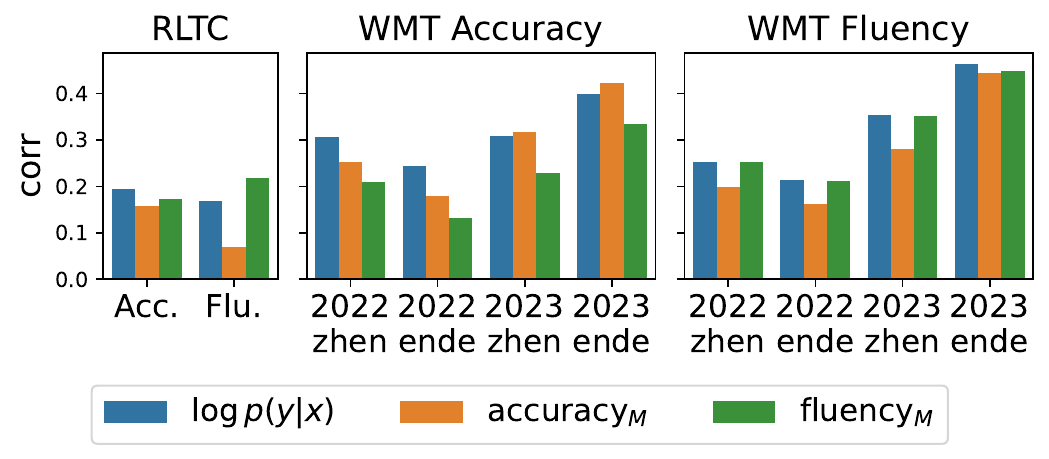}
    
    \caption{\accuracym\ and \fluencym\ predict human accuracy and fluency ratings for RLTC and WMT submissions to the general translation task in 2022 and 2023. \texttt{zhen} and \texttt{ende} refer to Chinese-English and English-German language pairs. All correlations reported are significant ($p<.001$).}
    \label{fig:acc-flu-alg}
\end{figure}

A similar conclusion is suggested by Figure~\ref{fig:acc-flu-alg}, which shows  Pearson correlations of translation probability (\p{y|x}; blue bars), \accuracym\ (\p{x|y}; brown bars) and \fluencym\ (\p{y}; green bars) with human ratings of accuracy and fluency for RLTC and MTMQM.\footnote{Values are in log scale and are ranked by percentile.} As expected, \accuracym\ shows a higher correlation with accuracy than fluency, and \fluencym\ shows the opposite pattern. Figure~\ref{fig:acc-flu-alg} however, suggests that \accuracym\ is not superior to \p{y|x} as a predictor of accuracy, and that \fluencym\ is not superior to \p{y|x} as a predictor of fluency. One reason why our model estimates of accuracy and fluency depart from human ratings is that \accuracym\ and \fluencym\ are sensitive to segment length. For example, a longer segment will have lower \fluencym\ than a shorter segment even if the two are both perfectly idiomatic.

\section{Conclusion}

We showed that accuracy and fluency and \p{x|y} and \p{y} both trade off when translating individual source segments.
This finding suggests that current protocols for assessing translation quality may need to be adjusted.
Human assessments for recent WMT General Tasks are performed using Direct Assessment and Scalar Quality Metrics (DA+SQM) \cite{kocmi2022findings,kocmi2023findings}. 
This approach  conflates meaning preservation and grammar into a single score indicative of overall quality of a translation.
In contrast, MQM is much more costly, but produces highly detailed  scores that use multiple sub-categories for both accuracy and fluency. Future approaches could therefore consider a middle ground that extends DA+SQM to include accuracy and fluency as independent aspects as in WMT16 \cite{bojar2016findings}.
This direction would allow automatic MT evaluation metrics such as BLEURT \cite{sellam2020bleurt} and COMET \cite{rei2022comet} (both fine-tuned to DA scores) to be adapted to provide independent scores for accuracy and fluency.

Our results also suggest the value of developing MT models that navigate the accuracy-fluency tradeoff in human-like ways. In some settings (e.g.\ translating legal texts) accuracy is more important than fluency \cite{popovic2020relations,martindale2018fluency,vela2015predicting,specia2011predicting,martindale2019identifying}, but in others (e.g.\ translating informal conversation) fluency may take priority \cite{poibeau2022human,frankenberg2022can}.  One natural approach to navigating the accuracy-fluency tradeoff builds on noisy channel models
\cite{yu2016neural,yee2019simple,muller2020domain}, which incorporate both \p{y} and \p{x|y} along with tradeoff parameters that specify the relative weights of the two. Tuning these parameters for specific registers may allow a model to find the right balance between accuracy and fluency in each case.

\section{Limitations} 
Although we provided evidence for both accuracy-fluency and \accuracym-\fluencym\ tradeoffs in translation, we did not explore semantic and grammatical features that may predict which source segments produce the greatest tradeoffs. Outside of our simulation we do not have access to ground-truth values of \p{x|y} and \p{y}, and are only able to approximate these values using specific NMT models.  Our work is also limited by the fact that MTMQM only includes translations generated by certain kinds of NMT models, and it is possible that our results do not generalize to translations generated by other types of models, such as statistical or rule-based MT systems. Finally, both RLTC and MTMQM have accuracy and fluency ratings derived from error annotations that are very similar in range. This constraint makes quality assessment and comparison at the segment level challenging.

\section*{Ethics Statement}
We do not foresee any potential risks and harmful use of our work. Our analyses are based on licensed data which are freely available for academic use.

\section*{Acknowledgements}
This work was supported by ARC FT190100200. 

\bibliography{anthology,custom}
\bibliographystyle{acl_natbib}

\appendix

\section{Appendix}
\label{sec:appendix}


\subsection{Data specification}
\subsubsection{Corpora}
The CRITT Translation Process Research Database \cite{carl2016critt} is a collection of translation behavioural data in the area of Translation Process Research. From the public CRITT database we obtain 15 studies across 13 pairs of languages: RUC17 (enzh, \citealp{carl2019machine}), ENJA15 (enja, \citealp{carl2016japanese}), NJ12 (enhi, \citealp{carl2016critt}), STC17 (enzh, \citealp{carl2019machine}), SG12 (ende, \citealp{nitzke2019problem}), ENDU20 (ennl, \citealp{vanroy2021syntactic}), BML12 (enes, \citealp{mesa2014gaze}), ACS08 (daen, \citealp{sjorup2013cognitive}), MS13 (ptzh, \citealp{schmaltz2016cohesive}), JLG10 (pten, \citealp{alves2013investigating}), BD13 (daen, \citealp{dragsted2010coordination}), LWB09 (daen, \citealp{jensen2009effects}), DG01 (plfr, \citealp{plonska2016problems}), BD08 (daen, \citealp{dragsted2010coordination}) and CREATIVE (enzh, \citealp{vieira2023translating}).\footnote{\url{https://sites.google.com/site/centretranslationinnovation/tpr-db/public-studies}} After deduplication and removing source segments with fewer than 4 unique translations, the total number of source segments included is 399, each with an average of 10.9 unique translations.

RLTC is a subset of the Russian Learner Translator Corpus that has been aligned at the segment level  by \citet{kunilovskaya2023translationese}. We include a total of 1079 source segments from 5 genres: `Essay', `Informational', `Speech', `Interview' and `Educational'. The average number of unique translations for each source segment is 10.5. 

MTMQM is obtained from \cite{freitag2021experts}, which contains translations of TED talks and news data from the test sets of WMT General Tasks between 2020 and 2023.\footnote{\url{https://github.com/google/wmt-mqm-human-evaluation}} The translations are annotated with MQM labels. After preprocessing we are left with 11219 source segments and an average of 9.9 unique translations per source segment.



\subsection{Alternative result with M2M100 translation model}
\begin{figure}[t]
         \includegraphics[width=\linewidth]{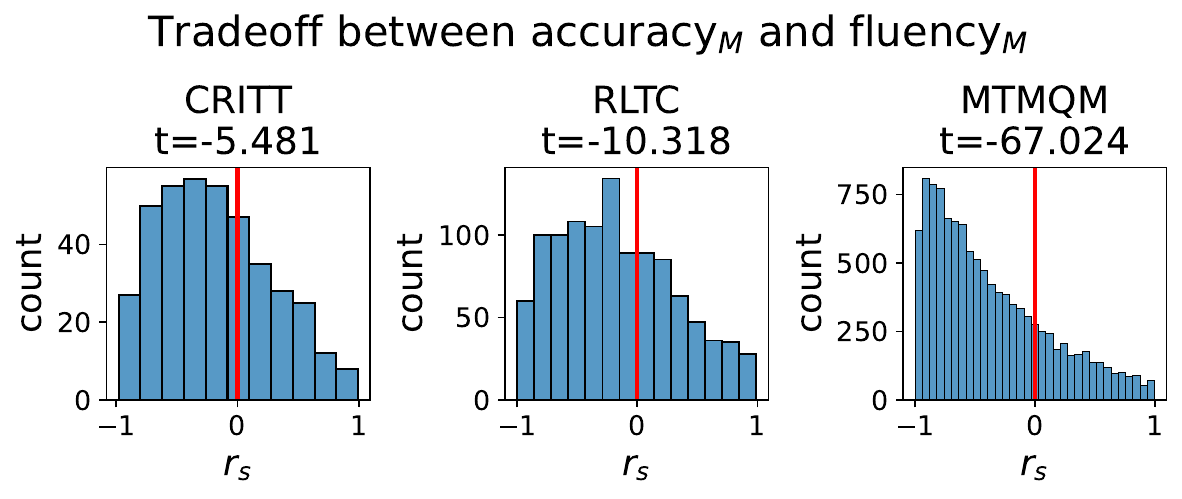}
         \caption{Histogram of tradeoffs between estimated \p{x|y} and \p{y} estimated by M2M100, which is analogous to Figure~\ref{fig:tradeoff} in the main text. When analyzed at the corpus level, the correlations $r_c$ for 
  for CRITT, RLTC and MTMQM are $.689$, $.703$ and $.801$ respectively ($p<.001$ in all cases).}
         \label{fig:tradeoff_m2m100}
\end{figure}

\begin{figure}[t]
    \includegraphics[width=\linewidth]{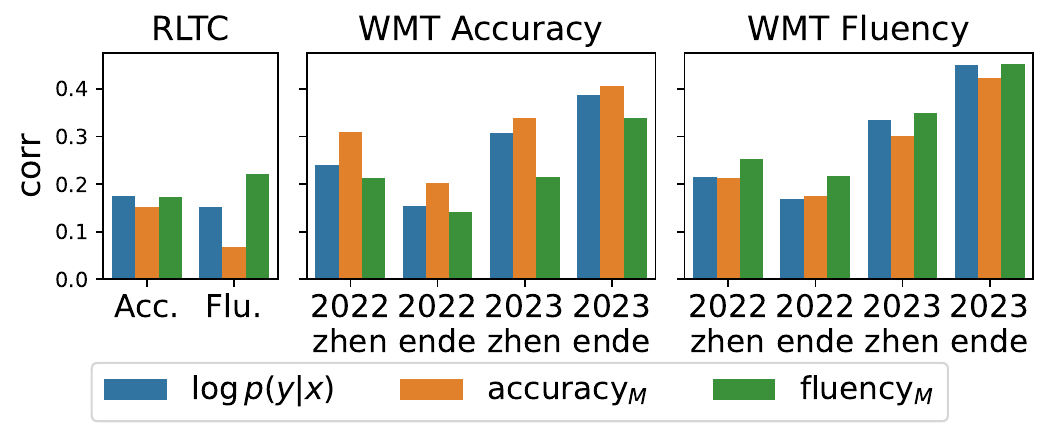}
    \caption{\accuracym\ and \fluencym\ estimates based on  M2M100 predict  human accuracy and fluency ratings ($p<.05$). The figure is analogous to Figure~\ref{fig:acc-flu-alg}.}
    \label{fig:acc-flu-alg-m2m100}
\end{figure}

In Figure~\ref{fig:tradeoff_m2m100} and \ref{fig:acc-flu-alg-m2m100}, we replicate our findings of \accuracym\ and \fluencym\ in Section~\ref{sec:prob_tradeoff} and \ref{sec:rating_tradeoff} with estimates based on M2M100 (1.2B variant) \cite{fan2021beyond}.\footnote{\url{https://huggingface.co/facebook/m2m100_1.2B}}

\subsection{Tradeoff examples}
Tables~\ref{tab:example1}, \ref{tab:example3} and \ref{tab:example2} include the full set of translations plotted in Figure~\ref{fig:simpson}.  The tables specify accuracy, fluency, \accuracym, \fluencym\ and translation probability \p{y|x} for each segment. All translations listed are submissions to the  WMT General Task between 2020 to 2022. 

\begin{table*}
\centering
\small
\begin{tabular}{l}
\toprule

Ich gab Ihnen eine Rückerstattung des Buches.\\ \{accuracy: 23.0, fluency: 25.0, \accuracym: -10.81, \fluencym: -56.0, $\log p(\bs{y}|\bs{x})$: -10.31\} \\\midrule
Ich habe dir eine Rückerstattung des Buches ausgestellt.\\ \{accuracy: 23.0, fluency: 25.0, \accuracym: -5.84, \fluencym: -62.5, $\log p(\bs{y}|\bs{x})$: -12.44\} \\\midrule
Ich habe dir das Buch zurückerstattet.\\ \{accuracy: 23.0, fluency: 25.0, \accuracym: -17.5, \fluencym: -44.25, $\log p(\bs{y}|\bs{x})$: -7.63\} \\\midrule
Ich habe Ihnen das Buch erstattet.\\ \{accuracy: 24.0, fluency: 25.0, \accuracym: -15.19, \fluencym: -43.25, $\log p(\bs{y}|\bs{x})$: -9.06\} \\\midrule
Ich habe Ihnen das Buch zurückerstattet.\\ \{accuracy: 24.2, fluency: 25.0, \accuracym: -17.25, \fluencym: -43.5, $\log p(\bs{y}|\bs{x})$: -7.28\} \\\midrule
Ich habe Ihnen eine Rückerstattung des Buches ausgestellt.\\ \{accuracy: 24.3, fluency: 24.67, \accuracym: -6.13, \fluencym: -64.0, $\log p(\bs{y}|\bs{x})$: -12.13\} \\\midrule
Ich stellte Ihnen eine Rückerstattung des Buches aus.\\ \{accuracy: 25.0, fluency: 23.0, \accuracym: -6.44, \fluencym: -70.0, $\log p(\bs{y}|\bs{x})$: -14.75\} \\\midrule
Ich habe Ihnen eine Rückerstattung für das Buch erteilt.\\ \{accuracy: 25.0, fluency: 24.0, \accuracym: -11.56, \fluencym: -63.0, $\log p(\bs{y}|\bs{x})$: -14.19\} \\
\bottomrule
\end{tabular}
\caption{Translations of \textit{I issued you a refund of the book.} (plotted in orange in Figure~\ref{fig:simpson}). Accuracy and fluency scores are derived from MQM ratings, and \accuracym\ and \fluencym\ are estimates of $\log$ \p{x|y} and $\log$ \p{y} derived from an NMT model. }
\label{tab:example1}

\end{table*}

{\small
\centering
\onecolumn
\begin{xltabular}{\textwidth}{X}
\toprule

Ashanti Development arbeitet seit fast 20 Jahren mit einer wachsenden Anzahl von Gemeinden in der Region Ashanti in Ghana zusammen und unterstützt sie in den Bereichen Wasser und sanitäre Einrichtungen, Bildung, Gesundheitsversorgung, Baumpflanzung und Landwirtschaft.\\ \{accuracy: 19.0, fluency: 25.0, \accuracym: -120.5, \fluencym: -498.0, $\log p(\bs{y}|\bs{x})$: -27.0\} \\\midrule
Ashanti Development arbeitet seit fast zwanzig Jahren mit einer ständig wachsenden Anzahl von Gemeinden in der Region Ashanti in Ghana zusammen, engagiert sich mit Gemeinden und unterstützt Wasser und Sanitärversorgung, Bildung, Gesundheitsversorgung, Baumpflanzung und Landwirtschaft. Gemeinschaften erlangen das Wissen, um ihre eigene Entwicklung einzubetten und zu unterstützen.\\ \{accuracy: 22.0, fluency: 24.0, \accuracym: -47.5, \fluencym: -748.0, $\log p(\bs{y}|\bs{x})$: -47.25\} \\\midrule
Ashanti Development arbeitet seit fast 20 Jahren mit einer ständig wachsenden Zahl von Gemeinden in der Ashanti-Region in Ghana zusammen, engagiert sich für Gemeinden und bietet Unterstützung in den Bereichen Wasser und sanitäre Einrichtungen, Bildung, Gesundheitsversorgung, Baumpflanzung und Landwirtschaft. Communities erwerben das Wissen, um ihre eigene Entwicklung zu verankern und zu unterstützen.\\ \{accuracy: 22.0, fluency: 25.0, \accuracym: -46.5, \fluencym: -832.0, $\log p(\bs{y}|\bs{x})$: -49.0\} \\\midrule
Ashanti Development arbeitet seit 20 Jahren mit einer immer größeren Zahl von Gemeinden in der Region Ashanti in Ghana zusammen, engagiert sich mit Gemeinden und unterstützt Wasser und Sanitärversorgung, Bildung, Gesundheitsversorgung, Baumpflanzung und Landwirtschaft.\\ \{accuracy: 23.0, fluency: 24.9, \accuracym: -101.0, \fluencym: -516.0, $\log p(\bs{y}|\bs{x})$: -39.25\} \\\midrule
Ashanti Development arbeitet seit fast 20 Jahren mit einer ständig wachsenden Anzahl von Gemeinden in der Ashanti-Region Ghanas zusammen, indem es sich mit Gemeinden beschäftigt und ihnen Unterstützung in den Bereichen Wasser und Sanitärversorgung, Bildung, Gesundheitsversorgung, Baumpflanzung und Landwirtschaft bietet.\\ \{accuracy: 23.0, fluency: 25.0, \accuracym: -98.5, \fluencym: -652.0, $\log p(\bs{y}|\bs{x})$: -29.625\} \\\midrule
Ashanti Development arbeitet seit fast 20 Jahren mit einer ständig wachsenden Zahl von Gemeinden in der Ashanti-Region in Ghana zusammen, arbeitet mit Gemeinden zusammen und unterstützt sie in den Bereichen Wasser und Abwasserentsorgung, Bildung, Gesundheitswesen, Baumpflanzung und Landwirtschaft. Gemeinschaften erwerben das Wissen, um ihre eigene Entwicklung zu verankern und zu unterstützen.\\ \{accuracy: 23.0, fluency: 24.0, \accuracym: -53.0, \fluencym: -828.0, $\log p(\bs{y}|\bs{x})$: -42.5\} \\\midrule
Ashanti Development arbeitet seit fast 20 Jahren mit einer ständig wachsenden Anzahl von Gemeinden in der Ashanti-Region in Ghana zusammen, engagiert sich für Gemeinden und unterstützt sie bei Wasser- und Sanitärversorgung, Bildung, Gesundheitswesen, Baumpflanzung und Landwirtschaft. Gemeinschaften gewinnen das Wissen, um ihre eigene Entwicklung einzubetten und zu unterstützen.\\ \{accuracy: 24.0, fluency: 23.0, \accuracym: -47.5, \fluencym: -784.0, $\log p(\bs{y}|\bs{x})$: -45.25\} \\\midrule
Ashanti Development arbeitet seit fast 20 Jahren mit einer stetig wachsenden Anzahl von Gemeinden in der Ashanti-Region in Ghana zusammen, engagiert sich mit Gemeinden und bietet Unterstützung in den Bereichen Wasserversorgung und Abwasserentsorgung, Bildung, Gesundheitsversorgung, Baumpflanzung und Landwirtschaft. Gemeinden erwerben das Wissen, um ihre eigene Entwicklung zu verankern und zu unterstützen.\\ \{accuracy: 24.0, fluency: 24.0, \accuracym: -49.5, \fluencym: -848.0, $\log p(\bs{y}|\bs{x})$: -42.25\} \\\midrule
Ashanti Development arbeitet seit fast 20 Jahren mit einer stetig wachsenden Anzahl von Gemeinschaften in der Ashanti-Region von Ghana zusammen, engagiert sich in den Gemeinschaften und bietet Unterstützung in den Bereichen Wasser und Sanitär, Bildung, Gesundheitswesen, Baumpflanzung und Landwirtschaft. Die Gemeinschaften erwerben das Wissen, um ihre eigene Entwicklung zu verankern und zu unterstützen.\\ \{accuracy: 25.0, fluency: 22.0, \accuracym: -50.25, \fluencym: -828.0, $\log p(\bs{y}|\bs{x})$: -43.0\} \\\midrule
Ashanti Development arbeitet seit fast 20 Jahren mit einer ständig wachsenden Zahl von Gemeinden in der Ashanti-Region in Ghana zusammen, engagiert sich für Gemeinden und leistet Unterstützung bei Wasser- und Sanitärversorgung, Bildung, Gesundheitsversorgung, Baumpflanzung und Landwirtschaft. Gemeinschaften erlangen das Wissen, um ihre eigene Entwicklung zu verankern und zu unterstützen.\\ \{accuracy: 25.0, fluency: 24.0, \accuracym: -45.75, \fluencym: -816.0, $\log p(\bs{y}|\bs{x})$: -45.0\} \\\midrule
Ashanti Development arbeitet seit fast 20 Jahren mit einer ständig wachsenden Zahl von Gemeinden in der Ashanti-Region in Ghana zusammen und unterstützt sie in den Bereichen Wasserversorgung und Abwasserentsorgung, Bildung, Gesundheitsversorgung, Baumpflanzung und Landwirtschaft. Die Gemeinden erlangen das Wissen, um ihre eigene Entwicklung zu fördern und zu unterstützen.\\ \{accuracy: 25.0, fluency: 24.0, \accuracym: -74.0, \fluencym: -768.0, $\log p(\bs{y}|\bs{x})$: -42.0\} \\\midrule
Ashanti Development arbeitet seit fast 20 Jahren mit einer ständig wachsenden Zahl von Gemeinden in der Ashanti-Region in Ghana zusammen, engagiert sich für Gemeinden und leistet Unterstützung bei Wasser- und Sanitärversorgung, Bildung, Gesundheitswesen, Baumpflanzung und Landwirtschaft. Gemeinschaften erwerben das Wissen, um ihre eigene Entwicklung zu verankern und zu unterstützen.\\ \{accuracy: 25.0, fluency: 24.0, \accuracym: -46.25, \fluencym: -812.0, $\log p(\bs{y}|\bs{x})$: -46.25\} \\

\bottomrule

\caption{Translations of \textit{Ashanti Development has been working with an ever-expanding number of communities in the Ashanti region of Ghana for approaching 20 years, engaging with communities and providing support with water and sanitation, education, healthcare, tree planting and farming. Communities gain the knowledge to embed and support their own development.} These translations are plotted in green in Figure~\ref{fig:simpson}.}
\label{tab:example3}
\end{xltabular}
\twocolumn
}

{\small
\centering
\onecolumn
\begin{xltabular}{\textwidth}{X}
\toprule

Interessanterweise war eine der 12 Galaxien in z66OD ein riesiges Objekt mit einem riesigen Gaskörper, bekannt als Himiko, das zuvor 2009 vom Subaru-Teleskop gefunden wurde. „Es ist vernünftig, einen Protohaufen in der Nähe eines massereichen Objekts wie Himiko zu finden. Wir sind jedoch überrascht zu sehen, dass Himiko nicht im Zentrum des Protohaufens lag, sondern am Rande 500 Millionen Lichtjahre vom Zentrum entfernt.“ Sagte Masami Ouchi, ein Teammitglied am Nationalen Astronomischen Observatorium von Japan und der Universität von Tokio, die Himiko im Jahr 2009 entdeckte, dass die Beziehung zwischen den Himiko und den Himiko-Klöstern noch immer nicht verstanden wird.\\ \{accuracy: 0.0, fluency: 22.9, \accuracym: -286.0, \fluencym: -1904.0, $\log p(\bs{y}|\bs{x})$: -139.0\} \\\midrule
"""""""Interessanterweise war eine der 12 Galaxien in z66OD ein riesiges Objekt mit einem riesigen Gaskörper, bekannt als Himiko, das zuvor vom Subaru-Teleskop im Jahr 2009 gefunden wurde. """"""""Es ist vernünftig, einen Protocluster in der Nähe eines massiven Objekts wie Himiko zu finden. Wir sind jedoch überrascht zu sehen, dass Himiko nicht im Zentrum des Protoclusters, sondern am Rand 500 Millionen Lichtjahre vom Zentrum entfernt war"""""""", sagte Masami Ouchi, ein Teammitglied am Nationalen Astronomischen Observatorium von Japan und der Universität von Tokio, der Himiko im Jahr 2009 entdeckte. Ironischerweise soll die mythologische Königin Himiko auch abgeschieden von ihrem Volk gelebt haben. Ouchi fährt fort: """"""""Es ist immer noch nicht verstanden, warum Himiko nicht im Zentrum liegt. Diese Ergebnisse werden ein Schlüssel für das Verständnis der Beziehung zwischen Haufen und massiven Galaxien sein"""""""\\ \{accuracy: 1.0, fluency: 23.4, \accuracym: -125.0, \fluencym: -2624.0, $\log p(\bs{y}|\bs{x})$: -103.5\} \\\midrule
"""""""Interessanterweise war eine der 12 Galaxien in z66OD ein riesiges Objekt mit einem riesigen Gaskörper, bekannt als Himiko, das zuvor vom Subaru-Teleskop im Jahr 2009 gefunden wurde. „Es ist vernünftig, einen Protocluster in der Nähe eines massiven Objekts, wie Himiko, zu finden. Allerdings sind wir überrascht zu sehen, dass Himiko nicht im Zentrum des Protoclusters, sondern am Rande 500 Millionen Lichtjahre vom Zentrum entfernt war.“, sagte Masami Ouchi, ein Teammitglied am Nationalen Astronomischen Observatorium von Japan und der Universität von Tokio, der Himiko im Jahr 2009 entdeckte. Ironischerweise soll die mythologische Königin Himiko auch abgeschieden von ihrem Volk gelebt haben. Ouchi fährt fort: """"""""Es ist immer noch nicht verstanden, warum Himiko nicht im Zentrum liegt. Diese Ergebnisse werden ein Schlüssel für das Verständnis der Beziehung zwischen Haufen und und massiven galaxien sein."""""""""""""""\\ \{accuracy: 6.0, fluency: 24.0, \accuracym: -121.0, \fluencym: -2688.0, $\log p(\bs{y}|\bs{x})$: -143.0\} \\\midrule
"""""""Interessanterweise war eine der 12 Galaxien in z66OD ein riesiges Objekt mit einem riesigen Gaskörper, bekannt als Himiko, das zuvor 2009 vom Subaru-Teleskop gefunden wurde. „Es ist vernünftig, einen Protocluster in der Nähe eines massiven Objekts wie Himiko zu finden. Wir sind jedoch überrascht zu sehen, dass sich Himiko nicht im Zentrum des Protoclusters befand, sondern am Rande 500 Millionen Lichtjahre vom Zentrum entfernt“, sagte Masami Ouchi, Teammitglied am National Astronomical Observatory of Japan und der Universität Tokio, der Himiko 2009 entdeckte. Ironischerweise soll die mythologische Königin Himiko auch abgeschieden von ihrem Volk gelebt haben. Ouchi fährt fort: """"""""Es ist immer noch nicht verstanden, warum Himiko nicht im Zentrum liegt. Diese Ergebnisse werden ein Schlüssel für das Verständnis der Beziehung zwischen Haufen und massiven Galaxien sein."""""""""""""""\\ \{accuracy: 6.0, fluency: 22.7, \accuracym: -126.0, \fluencym: -2592.0, $\log p(\bs{y}|\bs{x})$: -123.0\} \\\midrule
Interessanterweise war eine der 12 Galaxien in z66OD ein riesiges Objekt mit einem riesigen Gaskörper, bekannt als Himiko, das zuvor 2009 vom Subaru-Teleskop gefunden wurde. „Es ist vernünftig, einen Protocluster in der Nähe eines massiven Objekts wie Himiko zu finden. Wir sind jedoch überrascht zu sehen, dass Himiko sich nicht im Zentrum des Protoclusters befand, sondern am Rand 500 Millionen Lichtjahre vom Zentrum entfernt“, sagte Masami Ouchi, Teammitglied am National Astronomical Observatory of Japan und der Universität Tokio, der Himiko 2009 entdeckte. Ironischerweise soll die mythologische Königin Himiko auch abseits ihres Volkes im Kloster gelebt haben. Ouchi fährt fort: „Es ist immer noch nicht verstanden, warum Himiko sich nicht im Zentrum befindet. Diese Ergebnisse werden ein Schlüssel zum Verständnis der Beziehung zwischen Haufen und massiven Galaxien sein.“\\ \{accuracy: 9.0, fluency: 22.0, \accuracym: -131.0, \fluencym: -2512.0, $\log p(\bs{y}|\bs{x})$: -108.0\} \\\midrule
"""""""Interessanterweise war eine der 12 Galaxien in z66OD ein riesiges Objekt mit einem riesigen Gaskörper, bekannt als Himiko, das 2009 vom Subaru-Teleskop gefunden wurde. """"""""Es ist vernünftig, einen Protokluster in der Nähe eines massiven Objekts zu finden, wie z Himiko. Wir sind jedoch überrascht zu sehen, dass sich Himiko nicht in der Mitte des Protoklusters befand, sondern am Rand von 500 Millionen Lichtjahren vom Zentrum entfernt. """""""" sagte Masami Ouchi, ein Teammitglied des Nationalen Astronomischen Observatoriums Japans und der Universität Tokio, das Himiko 2009 entdeckte. Ironischerweise soll die mythologische Königin Himiko auch im Kloster von ihrem Volk gelebt haben. Ouchi fährt fort: """"""""Es ist immer noch nicht klar, warum Himiko nicht im Zentrum liegt. Diese Ergebnisse werden ein Schlüssel zum Verständnis der Beziehung zwischen Clustern und massiven Galaxien sein."""""""""""""""\\ \{accuracy: 13.0, fluency: 20.7, \accuracym: -132.0, \fluencym: -2688.0, $\log p(\bs{y}|\bs{x})$: -127.0\} \\\midrule
"""""""Interessanterweise war eine der 12 Galaxien in z66OD ein riesiges Objekt mit einem riesigen Gaskörper, bekannt als Himiko, das zuvor vom Subaru-Teleskop im Jahr 2009 gefunden wurde. """"""""Es ist vernünftig, einen Protocluster in der Nähe eines massiven Objekts wie Himiko zu finden. Wir sind jedoch überrascht zu sehen, dass Himiko nicht im Zentrum des Protoclusters lag, sondern am Rande 500 Millionen Lichtjahre vom Zentrum entfernt"""""""", sagte Masami Ouchi, Teammitglied am Nationalen Astronomischen Observatorium Japans und der Universität Tokio, der Himiko 2009 entdeckte. Ironischerweise soll auch die mythologische Königin Himiko von ihrem Volk abgeschottet gelebt haben. Ouchi fährt fort: """"""""Es ist immer noch nicht klar, warum Himiko nicht in der Mitte liegt. Diese Ergebnisse werden ein Schlüssel zum Verständnis der Beziehung zwischen Clustern und massiven Galaxien sein."""""""""""""""\\ \{accuracy: 16.0, fluency: 21.3, \accuracym: -122.5, \fluencym: -2624.0, $\log p(\bs{y}|\bs{x})$: -111.0\} \\

\bottomrule

\caption{Translations of \textit{"""Interestingly, one of the 12 galaxies in z66OD was a giant object with a huge body of gas, known as Himiko, which was found previously by the Subaru Telescope in 2009. """"""""It is reasonable to find a protocluster near a massive object, such as Himiko. However, we're surprised to see that Himiko was located not in the center of the protocluster, but on the edge 500 million light-years away from the center."""""""" said Masami Ouchi, a team member at the National Astronomical Observatory of Japan and the University of Tokyo, who discovered Himiko in 2009. Ironically, the mythological queen Himiko is also said to have lived cloistered away from her people. Ouchi continues, """"""""It is still not understood why Himiko is not located in the center. These results will be a key for understanding the relationship between clusters and massive galaxies."""""""""""} These translations are plotted in blue in Figure~\ref{fig:simpson}.}
\label{tab:example2}
\end{xltabular}
\twocolumn
}

\end{document}